\documentclass{article}
\usepackage{ijcai26}

\usepackage{times}
\usepackage{soul}
\usepackage{url}
\usepackage[hidelinks]{hyperref}
\usepackage[utf8]{inputenc}
\usepackage[small]{caption}
\usepackage{graphicx}
\usepackage{amsmath}
\usepackage{amsthm}
\usepackage{booktabs}
\usepackage{algorithm}
\usepackage{algorithmic}
\usepackage[switch]{lineno}

\usepackage{multirow}
\usepackage{enumitem} 

\title{StellarF: A Physics-Informed LoRA Framework for Stellar Flare Forecasting \\
with Historical \& Statistical Data
}

\author{
Tianyu Su$^1$\and
Zhiqiang Zou$^{1,2,3}$*\and
Qingyu Lu$^1$\and
Feng Zhang$^1$\and
Ali Luo$^{3,4,5}$\and
Xiao Kong$^4$\and
Min Li$^1$\\
\affiliations
$^1$School of Computer Science, Nanjing University of Posts and Telecommunications, Nanjing, Jiangsu, China\\
$^2$Jiangsu Key Laboratory of Big Data Security and Intelligent Processing, Nanjing, Jiangsu, China\\
$^3$University of Chinese Academy of Sciences, Nanjing, Jiangsu, China\\
$^4$CAS Key Laboratory of Optical Astronomy, National Astronomical Observatories, Beijing, China\\
$^5$School of Astronomy and Space Science, University of Chinese Academy of Sciences, Beijing, China\\
\emails
zouzq@njupt.edu.cn*
}

\begin{document}

\maketitle

\begin{abstract}
Stellar flare forecasting represents a critical frontier in astrophysics, offering profound insights into stellar activity mechanisms and exoplanetary habitability assessments. Yet the inherent unpredictability of flare activity, rooted in stellar diversity and evolutionary stages, underpins the field's core challenges: (1) sparse, incomplete, noisy lightcurve data from traditional observations; (2) ineffective multi-scale flare evolution capture via single representations; (3) poor physical interpretability in data-driven models lacking physics-informed priors. To address these challenges, we propose StellarF, a physics-informed framework synergizing general Al with astrophysical domain knowledge via three core components: a unified preprocessing pipeline for lightcurve refinement (missing-value imputation, temporal patch partitioning, adaptive sample filtering); a Low-Rank Adaptation (LoRA)-finetuned large language model (LLM) backbone enhanced by first-order difference augmentation, flare statistical information, and flare historical record modules for multimodal fusion instead of only simple representations; and a novel physics-informed loss embedding a minimum rising rate prior, appended to the cross-entropy loss, to align with flare physics. Extensive experiments on Kepler and TESS datasets show StellarF achieves state-of-the-art performance across key metrics, setting new benchmarks for flare forecasting. This work bridges general AI with astrophysics, offering a practical, physically interpretable paradigm for transient event forecasting in time-domain astronomy.
\end{abstract}

\section{Introduction}

\begin{figure}[ht] 
\vskip 0.2in
\begin{center}
\centerline{\includegraphics[width=\columnwidth]{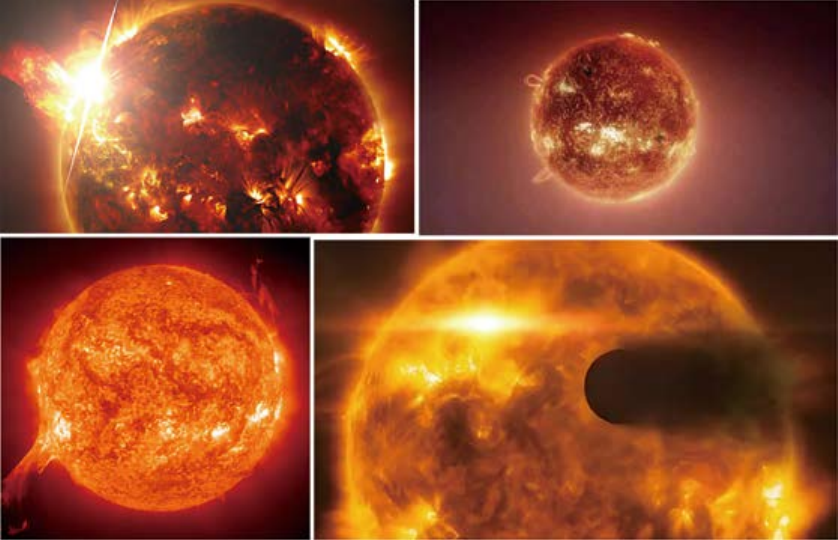}}
\caption{Some observed images of stellar flares.}
\label{fig1}
\end{center}
\vskip -0.2in
\end{figure}

Stellar flares are astronomical phenomena characterized by the rapid, intense and uncertain release of magnetic energy stored in stellar atmospheres, as shown in Figure \ref{fig1}. The high-energy radiation and particle fluxes emitted during flare eruptions not only profoundly influence the evolutionary pathways of their host stars, but also affect the atmospheric environments, magnetic structures, and even the habitability of nearby exoplanets \cite{segura2010effect,wang2025moving}. Within the broader context of astrophysical research, elucidating the eruption mechanisms of stellar flares is a critical step toward understanding stellar magnetic activity patterns, exploring complex star-planet system interactions, and defining potential habitable zones around exoplanets \cite{aschwanden2007solar,west2008constraining}. Paradoxically, despite the undeniable importance of flare forecasting, the diverse characteristics and evolutionary stages of stars result in highly variable and unpredictable flare activity patterns — a challenge that introduces significant uncertainties in both observational analysis and predictive modeling \cite{conroy2009propagation}, which further hinders high-accuracy forecasting efforts.

\begin{figure}[ht] 
\vskip 0.2in
\begin{center}
\centerline{\includegraphics[width=\columnwidth]{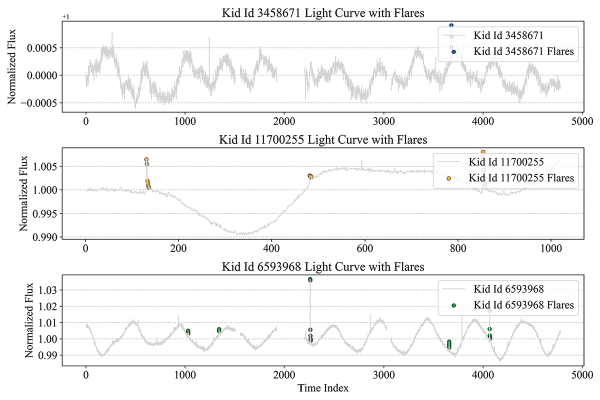}}
\caption{Light Curves of Three Stars with Flare Eruption Timings Marked. Visualizes observation period variations and light curve pattern differences across stars.}
\label{fig2}
\end{center}
\vskip -0.2in
\end{figure}

Achieving high-accuracy stellar flare forecasting faces significant challenges. First, traditional observations are constrained by instrumental limitations and temporal coverage, resulting in light curve data that commonly suffer from sparsity, incompleteness, and high noise \cite{dai2022flare7k}, as illustrated in Figure \ref{fig2}. These issues severely hinder the complete capture of dynamic flare characteristics, making it difficult to identify patterns and forecast future events. Second, existing studies predominantly rely on raw light curves as the sole single representation, failing to integrate critical information such as stellar physical properties and historical flare recurrence patterns. This leads to ineffective capture of multi-scale flare evolution dynamics. Finally, mainstream purely data-driven models lack prior constraints aligned with fundamental astrophysical principles. Consequently, their predictions often deviate from physical laws, exhibit low reliability, and struggle to extract features that simultaneously demonstrate statistical significance and physical interpretability from complex, noisy data. These limitations have long confined predictive accuracy to a performance bottleneck.
    
To systematically address the aforementioned core challenges, this study proposes StellarF, a multimodal framework for stellar flare forecasting. Unlike traditional analysis methods, StellarF addresses core challenges by synergizing general AI with astrophysical domain knowledge through three integrated components. First, it adopts a unified preprocessing pipeline for lightcurve refinement, which includes missing-value imputation, temporal patch partitioning, and adaptive sample filtering to leverage high-quality datasets. Second, the framework employs a Low-Rank Adaptation (LoRA)-finetuned\cite{hu2022lora} large language model (LLM) as its backbone enhanced by a multimodal fusion architecture integrating first-order difference augmentation, the flare statistical information and flare historical record to enhance its ability to identify multi-scale evolution patterns of stellar flares. Third, we introduce a novel physics-informed loss---embedded with a minimum rising rate prior and appended to the cross-entropy loss---to ensure consistency between model predictions and fundamental astrophysical principles. The key contributions of this study are summarized as follows:

\begin{itemize}
\item To advance deep cross-domain integration of general artificial intelligence techniques with astrophysical domain knowledge, we propose StellarF, a multimodal flare forecasting framework. StellarF realizes data augmentation via stacking raw stellar lightcurves with their first-order derivatives; to further enrich temporal feature representation, we specifically design two domain-specific modules---the flare statistical information Module and flare history record module---that encode the historical activity patterns and statistical features of stellar flares into structured natural language prompts. Leveraging a LLM lightweightly fine-tuned via LoRA, StellarF delivers physically consistent and interpretable predictions.

\item To further ensure the consistency between stellar flare prediction results and fundamental astrophysical principles, we propose a combined physics-informed cross-entropy loss function. This function builds on the standard cross-entropy loss and incorporates a physics-based prior constraint based on the minimum rising rate, which is calibrated on observed flare data without information leakage risks; its penalty term acts exclusively on positive samples (flare events), preserving end-to-end training differentiability while eliminating unphysical assumptions about non-flare lightcurves.

\item Extensive experiments on Kepler and TESS benchmark datasets demonstrate that StellarF outperforms state-of-the-art (SOTA) baselines, setting a new performance benchmark for stellar flare forecasting. This work offers a practical technical paradigm for AI-driven interdisciplinary research in astrophysics. Code and datasets are publicly available at \url{https://anonymous.4open.science/r/StellarFcast-E17A}.
\end{itemize}

\section{Literature Review}
\subsection{Traditional Methods}
In the early stages of stellar flare research, traditional approaches predominantly centered around physical models, statistical models, and empirical rule-based models. These methodologies laid a crucial foundation for subsequent studies, yet when confronted with the intricate system of stellar flare forecasting, their inherent limitations became evident.

\textbf{Statistical Models.} Time series analysis and regression analysis are commonly employed to explore the temporal dependencies of stellar luminosity and other physical quantities. For instance, the Autoregressive Integrated Moving Average (ARIMA) model \cite{box2015time} excels in capturing linear trends (e.g., solar sunspot number analysis), while the Generalized Autoregressive Conditional Heteroskedasticity (GARCH) \cite{daglis2024solar} model analyzes solar flares intensity fluctuations by modeling dynamic noise. However, these models struggle to handle nonlinear dynamics in complex flare activities.

\textbf{Physical Models.} Magnetohydrodynamic (MHD)-based models aim to explain stellar flare mechanisms via physical principles. A key example is the MHD instability model \cite{Kanya2020physics}, which simulated magnetic reconnection to predict major flares in Solar Cycle 24. While these models provide mechanistic insights, their high computational cost and strict parameter requirements limit application to diverse stellar types.

\textbf{Empirical Rule-based Models.} These models use historical stellar activity data and machine learning (e.g., decision trees, k-nearest neighbors (K-NN) algorithms) for prediction. Liu et al. \cite{liu2017predicting} trained a random forest model using vector magnetic data from Solar Dynamics Observatory/Helioseismic and Magnetic Imager (SDO/HMI) vector magnetic data, achieving moderate prediction accuracy for solar flares. However, limited by parameter accuracy and generalization, they struggle with flexible predictions in real-world scenarios.

Despite the contributions of traditional methods to the advancement of stellar flare forecasting, they generally fail to effectively capture complex nonlinear relationships and heavily rely on substantial prior knowledge. With the exponential growth of data volume and the continuous enhancement of computational capabilities, data-driven approaches have emerged as a new research trend. For example, Zhu et al.’s work \cite{zhu2025flare} integrates stellar properties and flare histories to improve light curve feature extraction and prediction accuracy, paving the way for new research directions.

\subsection{Time Series Analysis Methods}
Unlike the traditional methods, time series analysis methods play a pivotal role in numerous fields, especially in scenarios sensitive to time-dependent data scenarios, such as meteorological prediction and financial analysis. With the vigorous development of deep learning and big data technologies, this approach has achieved significant breakthroughs in handling complex patterns and long-term dependency relationships. Currently, mainstream time series analysis methods can be categorized into six types: MLPs, CNNs, RNNs, GNNs, Transformers, and PLMs.

\textbf{Multi-Layer Perceptrons (MLPs).} DLinear \cite{zeng2023transformers} designed a simple linear model that surpasses complex Transformer models in Long-Term Time Series Forecasting (LTSF) tasks. FITS \cite{xu2024fits} innovatively processes time series via interpolation operations in the complex frequency domain, attaining state-of-the-art performance with a lightweight architecture of only $\sim$10k parameters.

\textbf{Convolutional Neural Networks (CNNs).} CNNs effectively capture local temporal patterns through convolution operations, particularly when handling nonlinear and complex data. SCINet \cite{liu2022scinet} employs a recursive downsampling-convolution-interaction architecture, using multiple convolutional filters to derive valuable temporal features from downsampled subseries, significantly improving prediction accuracy.

\textbf{Recurrent Neural Networks (RNNs).} RNNs efficiently capture temporal dependencies, but suffer from the vanishing gradient problem. LSTM addresses this by introducing memory units, yet struggles with more complex temporal dependency patterns. DeepAR \cite{SALINAS20201181} proposes a method to generate accurate probabilistic predictions, training on large sets of related time series to estimate future probability distributions and reducing manual intervention effectively.

\textbf{Graph Neural Networks (GNNs).} GNNs \cite{li2018diffusion,wu2020connet} model time series variables as graph nodes and capture spatial dependencies between variables via graph convolutions. GCN \cite{kipf2017semisupervised} exhibits a linear scaling relationship with the number of graph edges, while learning hidden layer representations that encode local graph structures and node features, thereby markedly improving model performance.

\textbf{Transformer Models (Transformers).} Based on self-attention mechanisms, Transformers excel in capturing long-term dependencies and improving training efficiency, becoming a critical research direction in time series prediction. PatchTST \cite{nie2023patchtst} enhances long-sequence modeling efficiency via patch partitioning and local-global attention mechanisms. iTransformer \cite{liu2024itransformer} addresses the insufficient variable interaction modeling in traditional Transformers for time series prediction by embedding each variable's time series as independent tokens and using attention to capture multivariate correlations.

\textbf{Pre-trained Language Models (PLMs).} With the development of pre-trained language models (e.g., BERT \cite{devlin2019bert}, GPT series), increasing research applies them to time series analysis. Chronos \cite{ansari2024chronos} converts time series data into token sequences, leveraging language model architectures, pre-training on large-scale heterogeneous datasets, and data augmentation techniques to enhance generalization, demonstrating excellent zero-shot prediction performance.

Notwithstanding time series methods' remarkable achievements in general time series prediction, explorations in the astronomical domain---particularly for stellar flare forecasting---remain relatively scarce. To address this gap, we proposes StellarF, based on large language models (LLMs), that combines a unified preprocessing pipeline, multimodal feature representation enhancements, and physics-informed constraints. This provides new technical pathways for stellar flare forecasting research, with Section 3 detailing the framework's architecture.


\section{Methodology}

\begin{figure*}[ht] 
\vskip 0.2in
\begin{center}
\centerline{\includegraphics[width=0.98\linewidth]{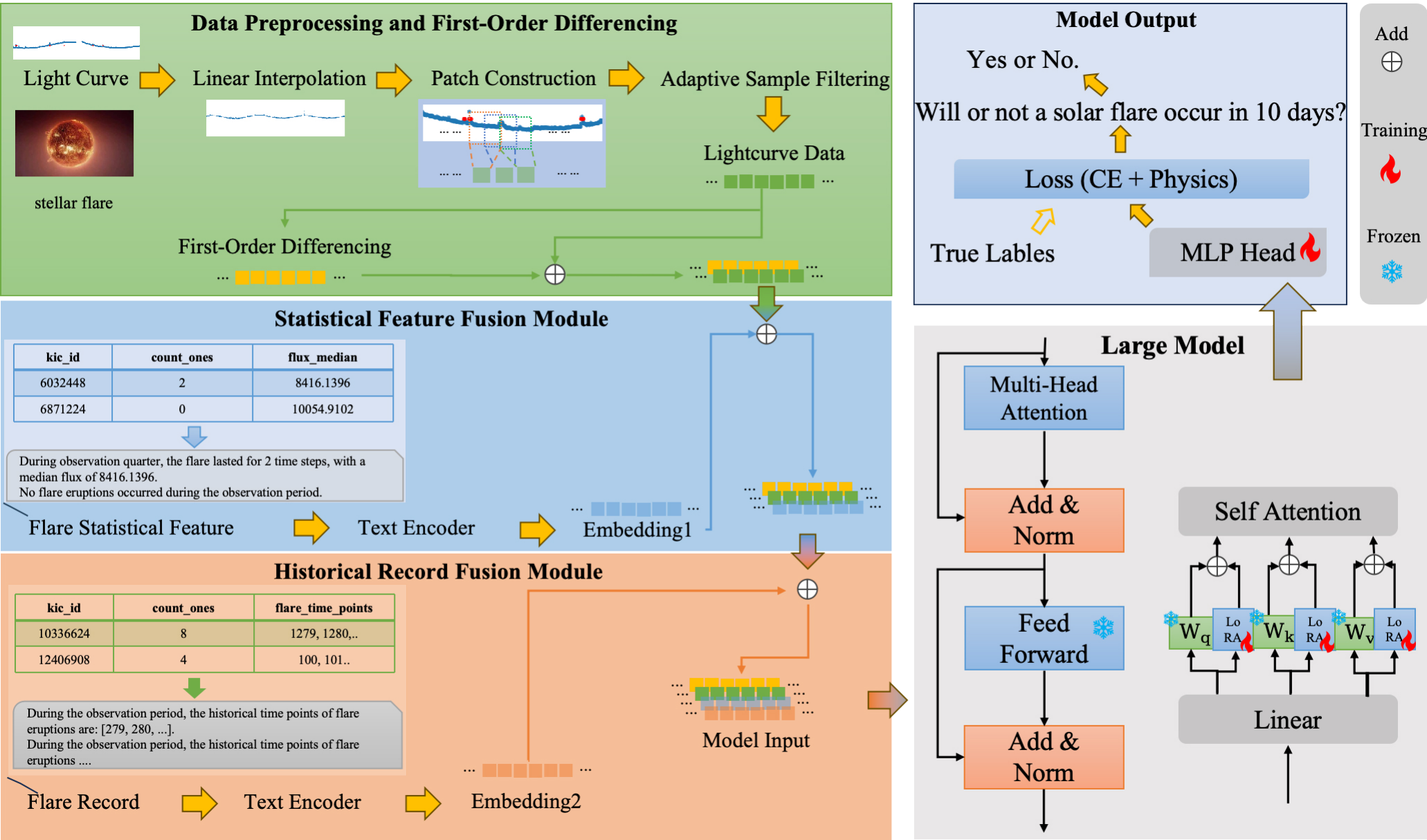}}
\caption{Overall Architecture of StellarF: It consists of a unified preprocessing pipeline (linear interpolation, patch-based sampling, adaptive sample filtering) for lightcurve refinement; a LoRA-tuned LLM backbone enhanced by first-order difference augmentation and dual multimodal fusion modules (statistical + historical record) for multimodal representation learning, followed by an MLP prediction head; and is further constrained by our novel physics-informed loss term, which embeds a minimum rising rate prior and is added to the cross-entropy (CE) loss to align with flare physics.}
\label{fig3}
\end{center}
\vskip -0.2in
\end{figure*}

In this section, we elaborate on the proposed StellarF model in detail, whose overall architecture is illustrated in Figure \ref{fig3}. Follwing sequential linear interpolation imputation, temporal patch reconstruction and adaptive sample filtering on the original dataset, we innovatively integrate the flare statistical features, historical flare event records and a differencing strategy into the light curve feature embedding process. This design constructs a multi-dimensional feature interaction mechanism to jointly capture the temporal evolution patterns, statistical laws and local transient features of stellar flares. Meanwhile, we specifically design a combined physics-informed cross-entropy loss function to enforce the consistency between model predictions and stellar physical laws. Finally, we fine-tune the LLM via LoRA, achieving efficient extraction of deep semantic features from multimodal data (i.e., light curve time series and textual information).

\subsection{Problem Definition}
Stellar flare forecasting is a scientific problem that involves making a prediction of whether a flare will occur within a specific future time window based on known observed light curve sequences. As the core observational data source, a stellar light curve systematically records the variation of a star's brightness over time. For a single star, the light curve can be represented as a time series $L^i=\{l_t^i\}_{t=1}^T$, where $l_t^i$ denotes the stellar photometric value measured for the i-th star at the time step $t$, and $T$ is the total number of observation time steps. 

To mitigate the non-stationarity of stellar light curve and enhance the model’s ability to capture the local transient luminosity variation features of flare eruptions, we introduce a first-order difference strategy for the original lightcurve time series. Its mathematical expression is given by: $D^i = \{\Delta d_t^i\}_{t=2}^T$, where $\Delta d_t^i = l_t^i - l_{t-1}^i$ ($\Delta d_t^i$ denotes the differential luminosity value of the i-th star at time step $t$). To align the differential sequence length with the original lightcurve, we pad the first position (corresponding to $t=1$) with 0. The differential feature sequence and original lightcurve together form a two-branch lightcurve feature representation.

Meanwhile, based on the known observed light curve sequence, historical flare information can be extracted, which can be represented as a set of time points $H^i=\{h_t^i\}_{t=1}^{n^i}$, where $h_t^i$ is the moment of the t-th flare eruption of the i-th star (marked by relative time within the observation period), and $n^i$ is the total number of flares in the historical records. Flare statistical information characterizes the overall features of stellar flare activity from a macro perspective, expressed as a tuple $S^i=\{n^i,m^i\}$, where $n^i$ represents the number of flares occurring during this observation period, and $m^i$ is the median flare flux. 

In summary, the input features for a single star can be integrated as $I^i=(L^i,D^i,H^i,S^i)$, where $L^i$ is the light curve time series, $D^i$ is the differencing information, $H^i$ is the historical flare information, and $S^i$ is the flare statistical information. The core objective of stellar flare forecasting is to construct a model $f(I^i;\theta)$ that, given input $I^i$ and model parameters $\theta$, can predict the probability $p^i$ of a flare occurring within the future time interval $[T+1, T+\Delta T]$ (where $\Delta T$ is the prediction range). The mathematical expression is as follows:
\begin{equation}
\mathrm{p}^\mathrm{i} = \mathrm{f}(I^i;\theta)
\end{equation}

\subsection{Light Curve Preprocessing and Embedding}
To mitigate the significant negative impact of missing values in flare light curves, we fill gaps with linear interpolation.

We design patch-based time series segmentation (tailored to Kepler’s 30-minute sampling and flare dynamics) with core hyperparameters:
\begin{itemize}
\item $patch\_len=512$ ($\approx$10.7-day context): covers full flare evolution and precursors;

\item $pred\_len=480$ ($\approx$10-day window): aligns with practical long-term flare warning (vs. short-term next-step prediction);

\item $stride=48$ ($\approx$1-day step):balances 90\% patch overlap (to reduce boundary-effect flare miss-detection) and training efficiency.
\end{itemize}
The segmented light curve can be expressed as $\hat{L} \in {R}^{N \times P}$, and $N=\left\lfloor {(K-P)/S} \right\rfloor$ is the number of patches with length $P$, where $K$ represents the total length of an original light curve and $S$ is the stride,  with corresponding differential features $\hat{D} \in {R}^{N \times P}$. 

Post-segmentation, we filter invalid samples (flat sequences derived from interpolated long-missing data) by thresholding the max 1st-difference (max\_rise) of flare samples (label=1): we remove samples with $max\_rise\leq0.001$ (normalized flux threshold, see Appendix \ref{app:preprocessing} for details.), yielding a high-quality dataset for subsequent model training.

\subsection{Textual Representation of Structured Data}
\textbf{Motivation} 
In the field of time-series prediction, numerous studies (e.g., Prompcast \cite{xue2024promcast}, LSTprompt \cite{liu-etal-2024-lstprompt}) have significantly improved prediction performance by integrating text prompts. Inspired by this, we investigate a critical question for stellar flare forecasting: how to deeply excavate and utilize the associated information of stellar light curves to boost prediction performance? Accordingly, we build upon the historical information module proposed from Zhu et al. and innovatively design a flare statistical information module. By fusing multi-dimensional stellar attribute information, this module strengthens the model's ability to capture flare occurrence patterns, yielding more effective feature representations and higher prediction accuracy.

\textbf{Historical Flare Information Module} 
The historical flare information module focuses on the local temporal features of individual stellar flare events. This module systematically collates the distribution of time points of historical flare events, providing the model with critical ``historical behavior'' reference. These time-series features not only reflect the periodic patterns of flare activity but also reveal the potential dynamic changes in stellar magnetic activity.

\textbf{Flare Statistical Information Module} 
The flare statistical information module, conversely, characterizes the global properties of stellar flare activity from a macro perspective. In the flare statistical information, all attribute value are expressed as precise scalar values. The unified encoding of semantic information of physical attributes and numerical features greatly assists the model in understanding the physical implications behind statistical laws.

\begin{figure}[ht]
\centering
\includegraphics[width=0.54\linewidth]{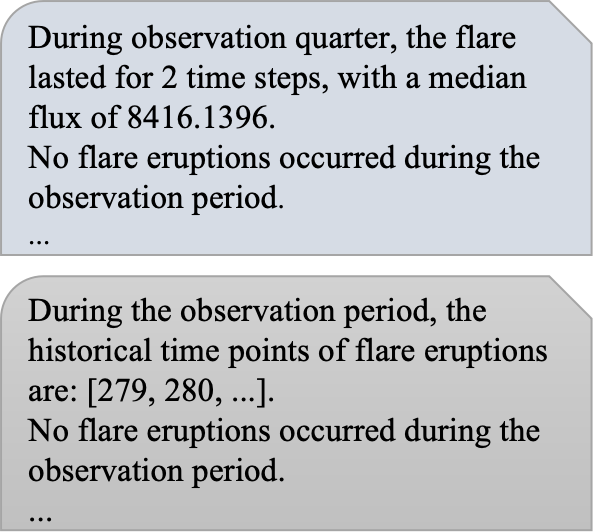} 
\caption{Textual Representations of Structured Data: Historical Flare Eruption Records (upper) and Statistical Summaries (count, median flux) (bottom) for Example Stellar Flares. Note: The model prompt used in the experiment was in Chinese.}
\label{fig8}
\end{figure}

The prompt templates for both modules are illustrated in Fig. \ref{fig8}. Both modules share a single text encoder for embedding all textual descriptions. Through the collaboration of dual modules, the model can simultaneously capture the temporal dynamics of stellar flares and statistical static features, thus constructing an input space with higher information completeness.

\subsection{Design of a Combined Physics-Informed Cross-Entropy Loss Function}

Traditional cross-entropy loss optimizes classification accuracy but ignores the core physical characteristic of stellar flares—significant transient flux rise—resulting in non-physical patterns that misalign with astrophysical dynamics. To address this, we design a physics-informed loss function that only applies physical prior constraints to positive flare samples ($label=1$). By penalizing positive samples that fail to satisfy flare physical properties, the loss guides the model to learn physically meaningful flare features and improves generalization.

The total loss is a weighted combination of cross-entropy classification loss and physical prior penalty loss for positive flares:
\begin{equation}
\mathcal{L}_{\text{total}} = \mathcal{L}_{\text{CE}} + \lambda_{\text{phys}} \cdot \mathcal{L}_{\text{phys}}
\end{equation}
where $\mathcal{L}_{\text{CE}}$ is the cross-entropy loss for binary classification, $\lambda_{\text{phys}}$ (default=0.1; additional experiments with $\lambda_{\text{phys}}$={0.3,0.6,0.9} are detailed in the Appendix \ref{app:loss_details}) balances the two loss components, and $\mathcal{L}_{\text{phys}}$ is the physical penalty loss (zero if no positive samples in the batch).

$\mathcal{L}_{\text{phys}}$ penalizes positive flares with maximum flux rise rate below a dataset-specific threshold rise\_threshold (derived via statistical analysis, see Appendix \ref{app:loss_details} for details.). Key hyperparameters include rise\_threshold (Kepler: 0.0175; TESS: 0.0132) and $\lambda_{\text{phys}}$ (Kepler and TESS: 0.1), with detailed calculations and distribution analysis Appendix \ref{app:loss_details}.

\subsection{Pre-trained Large Language Models}

Many studies have shown that training pre-trained language models (PLMs) from scratch often impairs model performance. However, by freezing most parameters and solely training a small subset of parameters, the powerful representation learning capabilities of PLMs can be effectively preserved. Specifically, we freeze most of the model’s parameters and introduce LoRA, allowing the large language model (LLM) to only fine-tune the learnable $Q$, $K$ and $V$ layers, ensuring robust adaptation while fully leveraging the prior knowledge of the pre-trained model. 

Thus, the model can effectively learn from historical flare records $\tilde{H}^i \in {R}^{T \times P}$ and flare statistical information $\tilde{S}^i \in {R}^{T \times P}$, where $T$ is the number of tokens in the segmented text description. Subsequently, we use a learnable linear projection layer to transform these features, mapping them to the same dimensional space as the lightcurve embedding $\hat{H}^i \in {R}^{N \times P}$ and $\hat{S}^i \in {R}^{N \times P}$, ensuring compatibility across different modal data. Based on this, we enhance the light curve embedding by incorporating differential information, flare statistical information, and flare historical records. Through this multi-modal feature fusion strategy, we finally construct an embedding representation $\hat{I}^i = (\hat{L}^i,\hat{D}^i,\hat{H}^i,\hat{S}^i) \in {R}^{N \times P \times 4}$, which effectively integrates spatial and temporal information of stellar flares.

\section{Experiments and Analysis}
\subsection{Experimental Setup}
\textbf{Datasets.} The Kepler Flare Dataset is a comprehensive compilation of stellar flare events observed by the Kepler Space Telescope. This dataset is constructed based on the flare event catalog proposed by \cite{yang2019flare} (a systematic Kepler Mission flare study). All light curves are long term time series data (approximately 29.4 minutes). It contains a total of 33,214 observational data entries from 3,420 stars across different observational quarters (Q1 to Q17). Each observational data entry covers a varying number of data points, from 1,021 to 4,780. For our framework’s temporal patch partitioning , we standardize each observation window to 512 data points, with the goal of predicting whether a flare event will take place in the upcoming 10 days (equivalent to 480 data points). The light curves of each stellar light curve are split into training, validation and test sets in an approximate ratio of 8:1:1 in chronological order, and the proportion of flare samples in the dataset is regulated to 50\% via random sampling.

\textbf{Baselines} 
We compare the proposed method with the following methods: DLinear, Informer \cite{zhou2021informer}, Autoformer \cite{wu2021autoformer}, iTransformer, MICN \cite{wang2023micn}, PatchTST, TimesNet \cite{wu2023timesnet} and GPT4TS \cite{zhou2023onefitall}---spanning MLP-based, RNN-based, Transformer-based, and PLM-based.    

\textbf{Evaluation Metrics} 
To evaluate the comprehensive performance of the model, we adopt the following evaluation metrics: Accuracy, Recall, Precision, F1-score (F1), and Area Under the ROC Curve (AUC). For the downstream task of stellar flare forecasting, we take Accuracy as the primary metric, with F1-score and AUC serving as auxiliary metrics for supplementary assessment.

\textbf{Implementation Details} 
We use the AdaW optimizer \cite{loshchilov2018decoupled} with the combined physics-informed cross-entropy loss function designed in this study, setting the learning rate to 1e-4, training for 200 epochs, and implementing an early stopping strategy (with a patience value of 10). For DLinear, TimesNet, and Transformer-based models, we refer to TSLib \cite{yu2024deep}, and GPT4TS is reproduced using the open-source code from the original paper. The text encoder employs BERT, while the pre-trained language model (PLM) adopts RoBERTa. All experimental datasets and the source code of our proposed model are publicly available in an open-source repository at: \url{https://anonymous.4open.science/r/StellarFcast-E17A}.

\subsection{Performance Comparison}

We compare StellarF with current state-of-the-art (SOTA) flare forecasting methods in Table \ref{table2} correspond to the mean and standard deviation across three runs. Additionally, we provide a comparison with Zhu et al.’s pioneering work to contextualize the field’s evolution, with detailed results in Appendix \ref{app:flare}.

Notably, by integrating four novel components—Flare Historical Records (FHR), Flare Statistical Information (FSI), First-order Differencing (FD), and a Physics-Informed Loss (PIL)---StellarF demonstrates exceptional performance across evaluation metrics: it achieves an accuracy of 62.67\% and an AUC of 68.86\%, outperforming all comparative models. From our analysis of these results, we derive three key insights:

\begin{table*}[ht]
\caption{Performance on the Kepler dataset with and without using Flare Historical Records (FHR), Flare Statistical Information (FSI), First-order Differencing, and the Physics-Informed Loss (PIL). Bold indicates the best. (\%)}
\label{table2}
\vskip 0.15in
\begin{center}
\begin{small}
\begin{sc}
\resizebox{0.98\linewidth}{!}{
\begin{tabular}{lcccccccccccc}
\toprule
                          & \multicolumn{5}{c}{LC}  & \multicolumn{5}{c}{LC + FHR + FSI + FD + PIL} \\ 
\multirow{-2}{*}{Methods} & {\textbf{Accuracy}} & {\textbf{Recall}} & {\textbf{Precision}} & {\textbf{F1}}  & {\textbf{AUC}} & {\textbf{Accuracy}}                             & {\textbf{Recall}}                               & { \textbf{Precision}}                            & { \textbf{F1}}                                   & { \textbf{AUC}}                                  \\ 
\midrule
DLinear                   & {48.10 ± 0.00}      & {99.90 ± 0.00}    & {48.10 ± 0.00}       & {64.93 ± 0.00} & {45.46 ± 0.00} & {52.55 ± 0.00}          & {89.40 ± 0.00}          & {50.38 ± 0.00}          & {64.44 ± 0.00}          & { 60.77 ± 0.00}          \\
Informer                  & {48.10 ± 0.00}      & {1.00 ± 0.00}     & {48.10 ± 0.00}       & {64.96 ± 0.00} & {53.60 ± 0.00} & {51.55 ± 4.88}          & {87.04 ± 18.33}         & {50.82 ± 3.85}          & {62.84 ± 3.00}          & {62.29 ± 0.02}          \\
Autoformer                & {50.73 ± 0.73}      & {67.74 ± 4.46}    & {49.09 ± 0.62}       & {56.88 ± 2.02} & {52.91 ± 1.39} & {50.73 ± 1.86}          & {92.79 ± 5.10}          & {49.43 ± 0.94}          & {64.42 ± 0.38}          & {59.79 ± 0.04}          \\
iTransformer              & {48.10 ± 0.00}      & {1.00 ± 0.00}     & {48.10 ± 0.00}       & {64.96 ± 0.00} & {51.70 ± 0.00} & {48.10 ± 0.00}          & {\textbf{1.00 ± 0.00}}           & {48.10± 0.00}           & {64.96 ± 0.00}          & {59.96 ± 0.00}          \\
MICN                      & {48.23 ± 0.02}      & {99.86 ± 0.20}    & {48.16 ± 0.01}       & {64.98 ± 0.06} & {50.71 ± 1.80} & {60.75 ± 0.00}          & {49.69 ± 0.00}          & {61.36 ± 0.00}          & {54.91 ± 0.00}          & {62.08 ± 0.00}          \\
PatchTST                  & {54.60 ± 0.00}      & {89.40 ± 0.00}    & {51.62 ± 0.00}       & {65.45 ± 0.00} & {61.90 ± 0.00} & {54.22 ± 4.36}          & {89.92 ± 7.32}          & {51.75 ± 2.80}          & {65.42 ± 0.31}          & {67.64 ± 0.23}          \\
TimesNet                  & {48.10 ± 0.00}      & {1.00 ± 0.00}     & {48.10 ± 0.00}       & {64.96 ± 0.00} & {53.44 ± 0.00} & {60.70 ± 0.00}          & {48.23 ± 0.00}          & {\textbf{61.70 ± 0.00}} & {54.14 ± 0.00}          & {62.44 ± 0.00}          \\
GPT4TS                    & {48.90 ± 0.00}      & {98.86 ± 0.00}    & {48.47 ± 0.00}       & {65.05 ± 0.00} & {56.23 ± 0.00} & {55.85 ± 0.00}          & {88.25 ± 0.00} & {52.44 ± 0.00}          & {\textbf{65.79 ± 0.00}} & {67.43 ± 0.00}          \\
StellarF(Ours)                  & -                   & -                 & -                    & -              & -              & { \textbf{62.70 ± 0.20}} & { 68.81 ± 0.10}        & {59.75 ± 0.23}          & {63.96 ± 0.09}          & {\textbf{68.87 ± 0.23}} \\ 
\bottomrule
\end{tabular}
}
\end{sc}
\end{small}
\end{center}
\vskip -0.1in
\end{table*}

\begin{enumerate}[label=(\arabic*)]
\item When using only raw light curve (LC) data, the PatchTST model also delivers strong performance, as can be seen from the left panel of the table. We hypothesize this is because PatchTST, by partitioning long-sequence data into patches, can deeply capture local statistical features and cross-scale long-range dependencies .

\item Our proposed StellarF model achieves the best overall performance, as can be seen from the right panel of the table. In contrast, models such as iTransformer and Informer exhibit subpar performance: iTransformer relies solely on a single attention mechanism to model temporal correlations, lacking targeted capture of the physical characteristics of stellar flares; Informer’s sparse attention mechanism suffers from an attention dilution issue in long-sequence flare data, failing to effectively focus on core features. StellarF, however, enables parameter-efficient lightweight fine-tuning via LoRA, and integrates four innovative components: FHR, FSI, PIL, and FD, ultimately achieving state-of-the-art prediction performance.

\item For all baseline models, incorporating FHR, FSI, FD, and PIL into their inputs leads to notable improvements in predictive performance. This finding underscores the critical role of these four components in stellar flare forecasting and validates the generalizable utility of such physics-informed priors and data augmentation strategies for enhancing time-series prediction model performance.
\end{enumerate}

\subsection{Ablation Study}
\textbf{Effectiveness of Each Module} 
To evaluate the effectiveness of each module in StellarF, we conduct an ablation study. The findings are presented in Table \ref{table3}, where we systematically remove one core module at a time to generate the following model variants: the full baseline model (StellarF), removal of the Flare Historical Records module (denoted as “StellarF w/o FHR”), removal of the Flare Statistical Information module (denoted as “StellarF w/o FSI”), removal of the First-order Differencing module (denoted as “StellarF w/o FD”), and removal of the Physics-Informed Loss module (denoted as “StellarF w/o PIL”).

\begin{table}[t]
\caption{The ablation analysis of StellarF. Bold indicates the best, and underlining denotes the second-best.}
\label{table3}
\vskip 0.15in
\begin{center}
\begin{small}
\begin{sc}
\resizebox{0.80\columnwidth}{!}{
\begin{tabular}{lccc}
\toprule
Methods              & {\textbf{Accuracy}} & {\textbf{F1}}  & {\textbf{AUC}} \\ 
\midrule
StellarF w/o FHR    & 62.30   & 62.89     & 66.57    \\
StellarF w/o FSI    & \underline{62.50}   & 63.27     & 67.85    \\
StellarF w/o FD      & 60.10  & 57.10     & 64.98  \\
StellarF w/o PIL     & 62.10   & \underline{65.04}     & \underline{68.53}   \\
StellarF w/o LoRA    & 59.95   & 64.42     & 67.05   \\
StellarF(Ours)             & \textbf{62.90}  & \textbf{65.26}     & \textbf{69.28}    \\ 
\bottomrule
\end{tabular}
}
\end{sc}
\end{small}
\end{center}
\vskip -0.1in
\end{table}

Experimental results validate the necessity of all five key components (four functional modules + LoRA) for StellarF's optimal performance, with FD being the most critical, FHR/FSI and LoRA providing distinct performance gains, and PIL offering modest incremental improvements. Additionally, key implementation choices (e.g., pre-trained language model selection, interpolation method) are verified via ablation studies, with detailed results in Appendix \ref{app:abla}.

\subsection{Integrated Gradients (IG) Attribution Analysis}

Interpretability is critical for validating ML model reliability in astrophysical applications. We conduct Integrated Gradients (IG) analysis to quantify light curve time-step contributions to StellarF’s binary prediction of future flare occurrence (yes/no)(as shown in Figure \ref{fig7}), with text/historical modalities zero-padded to isolate the light curve modality (details in Appendix \ref{app:ig_analysis}). For positive (flare) samples, the prominent peaks in absolute IG values, precisely aligned with pre-flare flux surges, confirm that StellarF prioritizes capturing predictive local flux variations for future flare forecasting. For negative (non-flare) samples, the model equally extracts discriminative local temporal features, with predictions driven by intrinsic sample patterns rather than randomness.

\begin{figure}[ht] 
\vskip 0.2in
\begin{center}
\centerline{\includegraphics[width=\columnwidth]{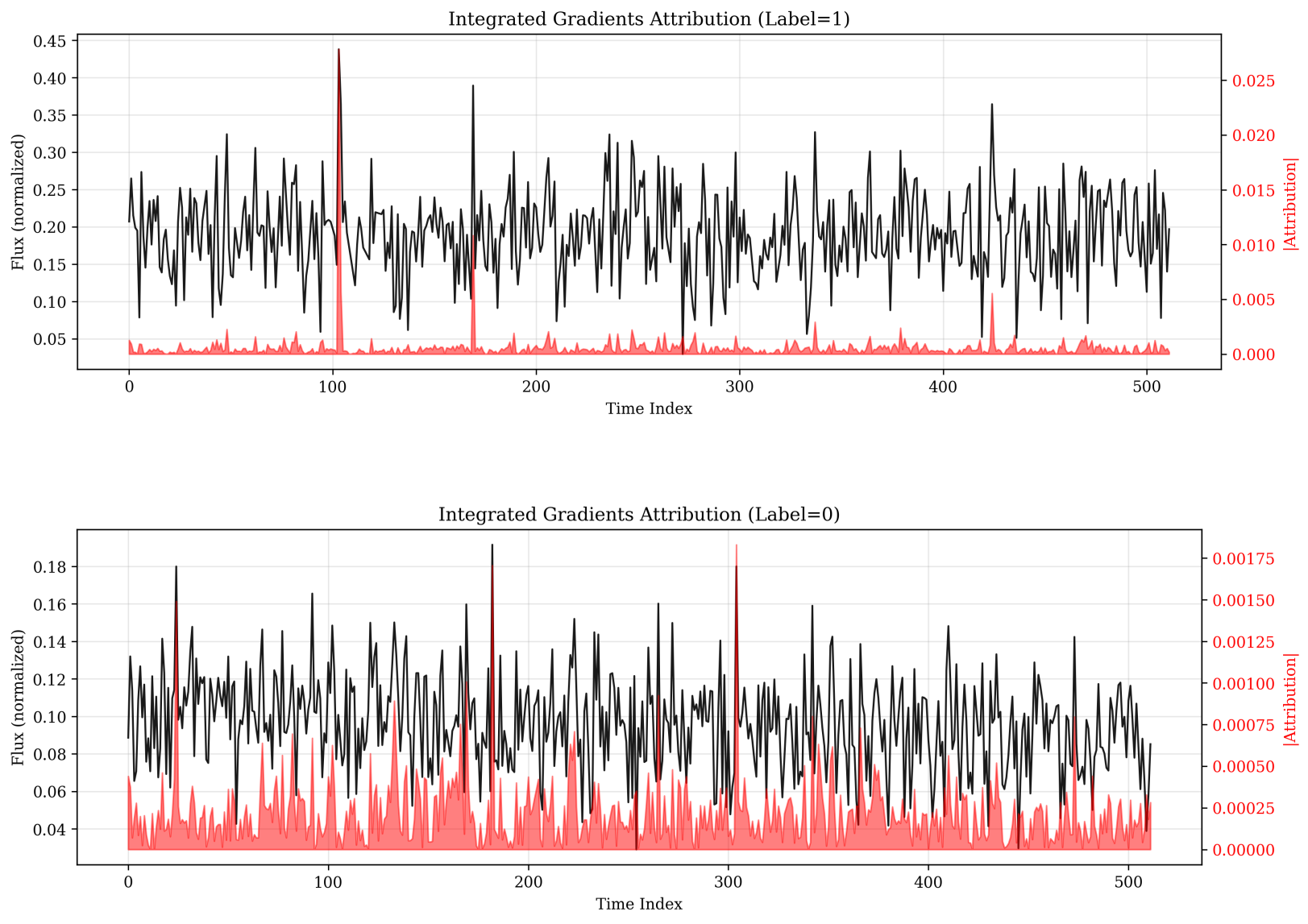}}
\caption{IG Attribution Maps (Upper: Positive Samples; Lower: Negative Samples)}
\label{fig7}
\end{center}
\vskip -0.2in
\end{figure}

We further validate StellarF’s generalization capability on an independent TESS dataset with heterogeneous observational characteristics. StellarF maintains state-of-the-art performance, confirming its adaptability to diverse flare forecasting scenarios—detailed results and analysis are presented in Appendix \ref{app:tess}.

\section{Conclusion}

In this paper, we propose StellarF, a multimodal forecasting framework integrating general AI techniques with astrophysical knowledge. StellarF introduces three key innovations: (1) a combined physics-informed cross-entropy loss embedding the minimum rising rate prior for physical consistency; (2) historical records and flare statistics (count and median flux) as multimodal inputs; and (3) a unified preprocessing pipeline enabling robust TESS generalization. Extensive experiments on Kepler and TESS datasets demonstrate that StellarF outperforms existing SOTA methods, establishing a new performance benchmark. It provides a practical and physically interpretable methodological paradigm for predicting astrophysical transients, advancing AI for Astrophysics as a critical next-generation technology.

\section*{Acknowledgments} 
We apperciate the contributions of Zhiqiang Zou, Ali Luo and Xiao Kong for their collaborative efforts. This work is supported by the National Natural Science Foundation of China under Grant 12473104.

\bibliographystyle{named}
\bibliography{stellarf}

\newpage
\appendix

\section{Supplementary Details on Stellar Light Curve Preprocessing}
\label{app:preprocessing}
\begin{figure}[ht] 
\begin{center}
\centerline{\includegraphics[width=\columnwidth]{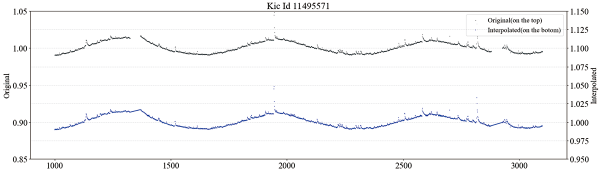}}
\caption{Linear Interpolation Effect on Light Curve Data. Original data (upper) with missing values vs. interpolated data (lower) preserving temporal trends.}
\label{fig4}
\end{center}
\end{figure}

To mitigate the significant negative impact of missing values in flare light curves, we fill gaps with linear interpolation (comparison in Fig. \ref{fig4}).

\begin{figure}[ht] 
\begin{center}
\centerline{\includegraphics[width=0.8\columnwidth]{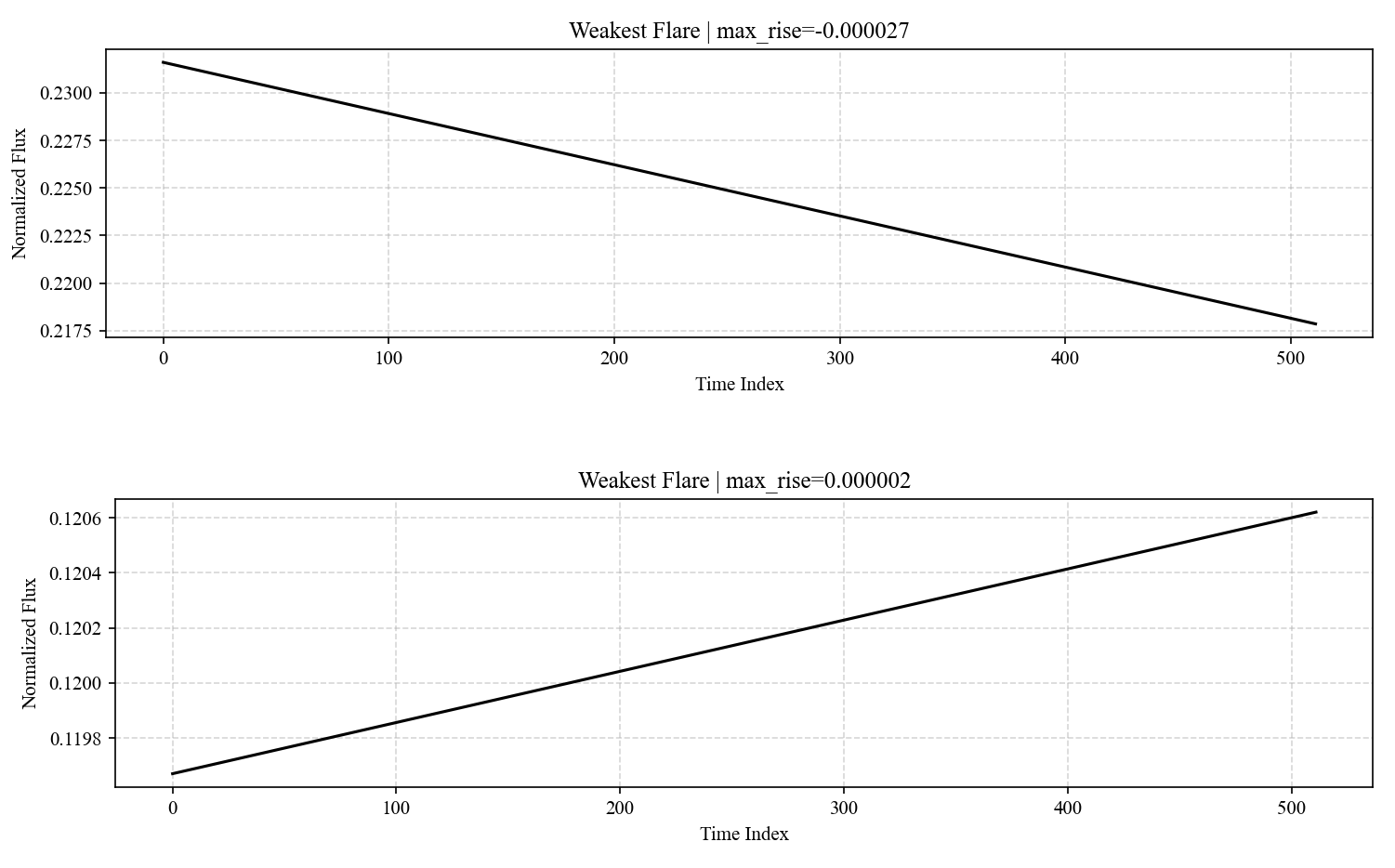}}
\caption{Anomalous Sample Screening via Max\_rise Threshold on Normalized Flux.}
\label{fig:anomaly_sample}
\end{center}
\end{figure}

\textbf{Filter Threshold Rationale}
The Kepler light curve data is normalized (eliminating baseline and instrumental biases). The threshold $\text{min\_rise}=0.001$ is used to distinguish valid flux variations from observational noise and interpolation errors. These flat sequences are derived from interpolated long-missing data and contain no valid flare features, as shown in Fig. \ref{fig:anomaly_sample}.

\section{Detailed Design of Physics-Informed Loss Function}
\label{app:loss_details}

\subsection{Physical Prior Penalty Loss Calculation}
The physical penalty loss $\mathcal{L}_{\text{phys}}$ is computed exclusively for positive flare samples (label=1) via four key steps:
\begin{enumerate}
    \item \textbf{First-order difference of light curves}:
          \begin{equation}
              \Delta l_{t}^i = l_{t}^i - l_{t-1}^i
          \end{equation}
          where $l_{t}^i$ denotes the normalized flux of the $i$-th sample at time step $t$.
    \item \textbf{Maximum flux rise rate per sample}:
          \begin{equation}
              \text{max\_rise}^i = \max_{t} (\Delta l_{t}^i)
          \end{equation}
    \item \textbf{Penalty for non-compliant samples}:
          \begin{equation}
              \text{penalty}^i = \text{ReLU}(\text{rise\_threshold} - \text{max\_rise}^i)
          \end{equation}
    \item \textbf{Batch-averaged penalty}:
          \begin{equation}
              \mathcal{L}_{\text{phys}} = \frac{1}{|B| + 10^{-8}} \sum_{i \in B} \text{penalty}^i \cdot label^i
          \end{equation}
          where $B$ denotes the batch. If no positive flares in the batch, $\mathcal{L}_{\text{phys}}=0$.
\end{enumerate}

\subsection{Key Hyperparameter Details}
\begin{figure}[t]
    \centering
    \includegraphics[width=1.0\linewidth]{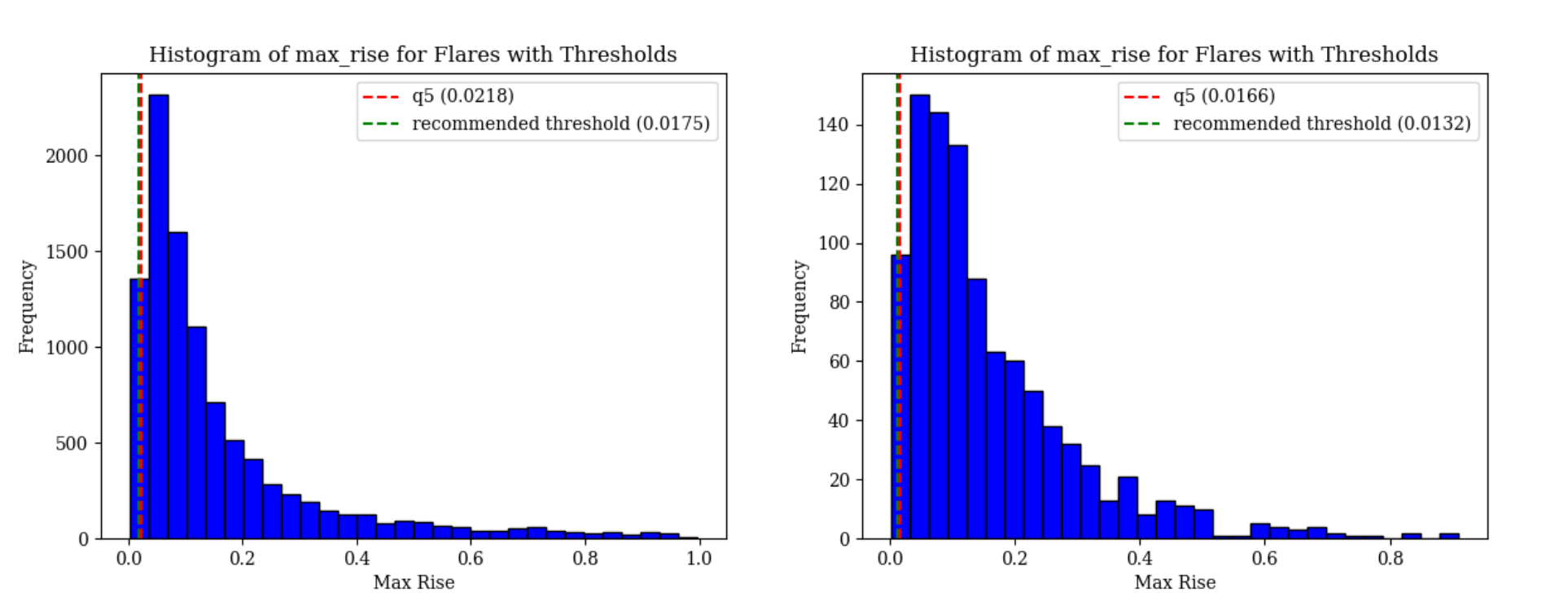} 
    \caption{Statistical distribution of maximum flux rise rate with corresponding thresholds. (left) Kepler dataset: dashed line marks the threshold $\text{rise\_threshold}=0.0175$. (right) TESS dataset: dashed line marks the threshold $\text{rise\_threshold}=0.0132$.}
    \label{fig:rise_threshold}
\end{figure}

\begin{itemize}
    \item $\text{rise\_threshold}$: Core physical prior threshold defining the minimum flux rise rate for valid stellar flares. Determined via dataset statistical analysis (Figure \ref{fig:rise_threshold}), with values 0.0175 (Kepler dataset) and 0.0132 (TESS dataset) to cover nearly all positive flare samples.
    \item $\lambda_{\text{phys}}$: Weight of the physical penalty loss (default value: 0.1, additional experiments with $\lambda_{\text{phys}} \in \{0.1, 0.3, 0.6, 0.9\}$ are shown in Table \ref{tab:ablation_lambda_phys}), used to adjust the strength of physical prior constraints on the model.
    \item $\text{conf\_threshold}$: Confidence threshold (default value: 0.5) for alternative penalty functions, reserved for future research extensions.
\end{itemize}

\begin{table}[h]
\caption{Ablation results of $\lambda_{\text{phys}}$ (0.1, 0.3, 0.6, 0.9) on StellarF's performance under identical experimental setups, where the default value 0.1 achieves optimal prediction accuracy. (\%)}
\label{tab:ablation_lambda_phys}
\vskip 0.15in
\begin{center}
\begin{small}
\begin{sc}
\resizebox{0.95\linewidth}{!}{
\begin{tabular}{lcccccc}
\toprule
                                        & \multicolumn{3}{c}{Kepler}  & \multicolumn{3}{c}{TESS} \\ 
\multirow{-2}{*}{Methods}              & {\textbf{Accuracy}} & {\textbf{F1}}  & {\textbf{AUC}} & {\textbf{Accuracy}}   & { \textbf{F1}}    & { \textbf{AUC}}    \\ 
\midrule
StellarF(0.3)    & {62.35}            & {63.50}          & {68.50}                  & {66.50}            & {71.92}          & {75.09}            \\
StellarF(0.6)    & {62.55}            & {64.52}          & {68.42}                  & {67.30}            & \textbf{72.15}          & {74.92}     \\
StellarF(0.9)    & {61.35}            & \textbf{65.69}   & {68.76}                  & {66.65}            & {72.06}          & {75.07}            \\ 
StellarF(0.1)(Ours)    & \textbf{62.90}     & {65.26}          & \textbf{69.28}     & \textbf{69.35}     & {71.31}          & \textbf{78.50}     \\
\bottomrule
\end{tabular}
}
\end{sc}
\end{small}
\end{center}
\vskip -0.1in
\end{table}

\section{Comparison with Foundational Pioneering Work}
\label{app:flare}
Table \ref{table: compare with zhu} presents core metric comparisons between our reproduced Zhu et al.’s (FLARE) method and StellarF under two prediction length settings: pred\_len=480 (our long-term prediction task) and pred\_len=48 (consistent with FLARE’s hyperparameters). In both setups, StellarF outperforms the reproduced FLARE method, while the latter only reaches 49.45\% in core Accuracy—failing to match the accuracy reported in the original FLARE paper.  We conjecture that this performance gap stems from two key limitations: first, critical stellar physical property data selected in their study is not available to us, which forced the omission of their original Stellar Physical Properties module; second, the complete reproducible code for their method remains unavailable due to non-disclosure, hindering our ability to replicate their experimental setup exactly. For consistency, we adopted our standardized data processing pipeline for this reproduction. Despite these constraints, the significant performance advantage of StellarF fully demonstrates the effectiveness of our four proposed novel components (FHR, FSI, FD, PIL). Our code and datasets are publicly available at \url{https://anonymous.4open.science/r/StellarFcast-E17A}.

For future work, we will make deliberate efforts to collect the critical stellar physical property data that was missing in this reproduction. With these data in hand, we will conduct more accurate and rigorous replication experiments of the original model, aiming to achieve a more fair and comprehensive performance comparison between the two methods.

\begin{table}[h]
\caption{Comparison between the reproduced version of Zhu’s method and our StellarF. Bold indicates the best. (\%)}
\label{table: compare with zhu}
\begin{center}
\begin{small}
\begin{sc}
\resizebox{0.90\columnwidth}{!}{
\begin{tabular}{lcccccc}
\toprule
& \multicolumn{3}{c}{pred\_len=480}  & \multicolumn{3}{c}{pred\_len=48} \\ 
\multirow{-2}{*}{Methods}              & {\textbf{Accuracy}} & {\textbf{F1}}  & {\textbf{AUC}} & {\textbf{Accuracy}} & {\textbf{F1}}  & {\textbf{AUC}}\\ 
\midrule
FLare    & 48.10   & 64.96     & 51.09    & 49.45   & \textbf{66.18}     & 50.56\\
StellarF(Ours)             & \textbf{62.90}  & \textbf{65.26}    & \textbf{69.28}   & \textbf{63.60}  & 60.99    & \textbf{66.80} \\ 
\bottomrule
\end{tabular}
}
\end{sc}
\end{small}
\end{center}
\end{table}

\section{Additional Ablation Studies}
\label{app:abla}
\subsection{Ablation Studies on PLM}
Table \ref{table: aba on plm} quantifies the impact of backbone Pre-trained Language Model (PLM) selection on StellarF’s predictive performance. StellarF achieves the poorest performance with DeBERTa, a deficiency we attribute to its architectural mismatch with flare data: its disentangled attention and document-level pre-training prioritize global semantic coherence (tailored for NLP) over local temporal feature detection, critical for identifying transient, sparse flares in long noisy time series. This impairs its ability to capture rare flare events—the core inference target.

\begin{table}[h]
\caption{The ablation analysis of PLM in StellarF. Bold indicates the best. (\%)}
\label{table: aba on plm}
\begin{center}
\begin{small}
\begin{sc}
\resizebox{0.70\columnwidth}{!}{
\begin{tabular}{lccc}
\toprule
Methods              & {\textbf{Accuracy}} & {\textbf{F1}}  & {\textbf{AUC}} \\ 
\midrule
BERT    & 61.15   & 60.22     & 65.70    \\
GPT2    & 60.35   & 54.08     & 63.09    \\
DeBERTa      & 48.10  & 64.96     & 53.27  \\
RoBERTa     & \textbf{62.90}  & \textbf{65.26}    & \textbf{69.28}   \\
\bottomrule
\end{tabular}
}
\end{sc}
\end{small}
\end{center}
\end{table}

Conversely, GPT-2 or BERT yield consistent performance gains across metrics, demonstrating better compatibility with flare time-series characteristics. RoBERTa delivers optimal performance across all core metrics, owing to two key advantages: (1) its bidirectional self-attention (inherited from BERT) outperforms GPT-2’s causal unidirectional attention for modeling light curve sequential dependencies; (2) its iterative refinements (dynamic masking, extended continuous pre-training) enhance sensitivity to subtle temporal patterns and faint flare signatures.

These results highlight the critical, non-trivial importance of deliberate PLM selection for optimizing stellar flare forecasting performance.

\subsection{Interpolation Method Ablation}
To address continuous missing values in stellar light curves, we ablate three common interpolation methods: linear, KNN, and periodic interpolation. As shown in Table \ref{table: aba on interpol}, linear interpolation outperforms the others across all core metrics. This superiority aligns with light curves’ intrinsic temporal properties: non-flare periods exhibit stable baseline flux with weak fluctuations, approximating linear trends—linear interpolation fits this baseline accurately without artificial deviations.

\begin{table}[ht]
\caption{Ablation analysis of three interpolation methods (linear, KNN, periodic) for missing value processing.}
\label{table: aba on interpol}
\begin{center}
\begin{small}
\begin{sc}
\resizebox{0.85\columnwidth}{!}{
\begin{tabular}{lccc}
\toprule
Methods              & {\textbf{Accuracy}} & {\textbf{F1}}  & {\textbf{AUC}} \\ 
\midrule
StellarF(periodic)    & 60.75   & 57.54     & 62.57    \\
StellarF(KNN)    & 62.05   & \textbf{66.52}     & 67.93    \\
StellarF(linear)(Ours)      & \textbf{62.90}  & {65.26}    & \textbf{69.28}   \\
\bottomrule
\end{tabular}
}
\end{sc}
\end{small}
\end{center}
\end{table}

For flare periods, flares are transient, low-proportion events, and continuous missing values rarely occur due to high time-resolution observations. Even for occasional gaps, linear interpolation avoids spurious signals and noise, unlike KNN interpolation (which propagates noise from adjacent non-flare time steps) or periodic interpolation (whose cycle constraints fail for non-periodic flares). Thus, linear interpolation is adopted as the standard missing value processing method, balancing accuracy and robustness to light curves’ intrinsic characteristics.


We evaluate the overall performance of StellarF against mainstream baseline models, and further complement the comparison with Zhu et al.’s pioneering work—the first to propose the core methodological framework for stellar flare forecasting in this domain. Given that the original code and full data processing details (e.g., sampling strategies, missing value imputation) of Zhu et al are not publicly available, we implement a reasonable reproduction of their approach based on the methodological descriptions in their publication. Two key limitations of this reproduction should be noted: (1) omission of the Stellar Physical Properties module (a core component of their framework) due to the unavailability of their mentioned dataset and associated feature engineering details; (2) adoption of our consistent data processing pipelines for fair comparison, as their specific processing protocols were not disclosed.

\section{Details of Integrated Gradients (IG) Attribution Analysis}
\label{app:ig_analysis}
This appendix provides complete technical details of the Integrated Gradients (IG) attribution analysis (omitted from the main text for brevity), including reproducible calculation parameters and standardized visualization protocols.

\subsection{IG Calculation Parameters}
IG attribution values for individual light curve samples were computed with the following fixed core parameters to ensure experimental reproducibility:
\begin{enumerate}
    \item \textbf{Baseline Selection}: A zero baseline (baseline\_type="zeros") was adopted as the reference to quantify the feature contribution of each time step.
    \item \textbf{Interpolation Steps}: 100 linear interpolation steps ($n_{\text{steps}}=100$) were set to balance calculation accuracy and computational efficiency.
    \item \textbf{Target Class Definition}: Consistent with the binary flare forecasting task, target classes were defined as 1 (flare, positive sample) and 0 (non-flare, negative sample).
\end{enumerate}

\subsection{Visualization Protocol}
A dual-axis plotting strategy was employed to visualize the IG attribution results and their correlation with light curve features:
\begin{itemize}
    \item \textbf{Primary Axis}: Displays the normalized flux of the stellar light curve (black line), representing the original temporal signal of the star.
    \item \textbf{Secondary Axis}: Shows the absolute IG attribution values (red filled area), quantifying the magnitude of feature importance for each time step.
\end{itemize}
Four high-resolution (300 dpi) visualization plots were generated, including flare (positive) samples and non-flare (negative) samples. These samples were randomly selected from the high-quality processed dataset to ensure representativeness across different flare intensities and light curve noise levels.

\section{Generalization Validation on TESS Dataset}
\label{app:tess}
To validate the generalization capability of the StellarF model, we introduce an independent validation dataset constructed from the light curves of the Transiting Exoplanet Survey Satellite (TESS) \cite{ricker2016transiting}. This dataset was processed in strict accordance with the identical data preprocessing pipeline for the Kepler dataset (Section 3.2 for details), including a unified sampling strategy, missing value imputation method, and flare event labeling criteria. Notably, TESS differs from Kepler in core observational characteristics: a shorter observational baseline (27 days per sector vs. Kepler’s 4+ years) and broader sky coverage. These differences make this validation a stringent test of StellarF’s adaptability to heterogeneous observational conditions.

For the TESS dataset, we adopted the identical experimental setup and comparative baselines as those used for the Kepler dataset, which include mainstream time-series forecasting models (PatchTST, iTransformer, Informer) and the reproduced pioneering work in this field (Section 4.2 for details). As presented in Table \ref{table:tess}, StellarF achieves the optimal overall performance across all core evaluation metrics, demonstrating that the model has robust generalization capability on the TESS dataset.

\begin{table}[h]
\caption{Performance of our updates on the TESS dataset. Bold indicates the best, and underlining denotes the second-best. (\%)}
\label{table:tess}
\begin{center}
\begin{small}
\begin{sc}
\resizebox{0.70\columnwidth}{!}{
\begin{tabular}{lccc}
\toprule
Methods              & {\textbf{Accuracy}} & {\textbf{F1}}  & {\textbf{AUC}} \\ 
\midrule
DLinear    & \underline{66.30}   & 70.12     & 73.95    \\
Informer    & 65.65   & \textbf{71.58}     & \underline{74.31}    \\
Autoformer     & 64.55  & 70.93     & 72.56  \\
iTransformer     & 48.80   & 65.59     & 72.58   \\
MICN    & 60.75   & 54.91     & 62.08   \\
PatchTST    & 53.80   & 67.44     & 72.90   \\
TimesNet    & 65.25   & 71.13     & 74.25   \\
GPT4TS    & 63.30   & 71.06     & 72.22   \\
Flare    & 48.80   & 65.59     & 50.00   \\
StellarF(Ours)             & \textbf{69.35}  & \underline{71.31}     & \textbf{78.50}    \\ 
\bottomrule
\end{tabular}
}
\end{sc}
\end{small}
\end{center}
\end{table}

\section{Reproducibility Checklist}
\label{app:reproducibility}

To ensure full reproducibility, we provide all code, data, and detailed instructions in a public repository: \url{https://anonymous.4open.science/r/StellarFcast-E17A}

The repository includes:
\begin{itemize}
    \item The complete implementation of the proposed algorithm;
    \item All datasets used in the experiments (including preprocessing scripts for external datasets);
    \item Configuration files specifying the exact hyperparameters for each experiment reported in the paper;
    \item A \texttt{requirements.txt} (or \texttt{environment.yml}) file listing all dependencies;
    \item A \texttt{README.md} with step-by-step instructions to reproduce every major result in under 30 minutes on a standard GPU.
\end{itemize}

All external datasets are properly cited and publicly available. No proprietary or restricted resources are used. We believe this setup enables any researcher to reproduce our results with minimal effort.

\end{document}